\documentclass[11pt,a4paper]{article}

\usepackage[hyperref]{emnlp2020}

\usepackage{times}
\usepackage{latexsym}

\usepackage{microtype}
\usepackage{tikz}
\usepackage{tabularx}

\usepackage{amsmath}
\usepackage{amsthm}
\usepackage{amssymb}

\usepackage{paralist}
\usepackage[capitalise]{cleveref}
\usepackage{comment}

\usepackage{booktabs}

\usepackage{url}

\DeclareMathOperator{\encoder}{encoder}
\DeclareMathOperator{\params}{\theta}
\DeclareMathOperator{\MLP}{MLP}

\aclfinalcopy 

\newcounter{trCounter}
\newif\iftrvar
\trvartrue
\iftrvar
\newcommand{\tim}[1]{{\small \color{red} \refstepcounter{trCounter}\textsf{[TR]$_{\arabic{trCounter}}$:{#1}}}}

\else
\newcommand{\tim}[1]{}

\fi

\newcommand{\ie}{\textit{i.e.}}

\title{Avoiding the Hypothesis-Only Bias in Natural Language Inference via Ensemble Adversarial Training}

\author{Joe Stacey\\
  University College London \\
  \texttt{ucakjd0@ucl.ac.uk} \\ \And
  Pasquale Minervini \\
  University College London \\
  \texttt{p.minervini@ucl.ac.uk} \\ \And
  Haim Dubossarsky \\
  University of Cambridge \\
  \texttt{hd423@cam.ac.uk} \\ \AND
  Sebastian Riedel \\
  University College London \\
  \texttt{s.riedel@cs.ucl.ac.uk} \\ \And
  Tim Rockt{\"a}schel \\
  University College London \\
  \texttt{t.rocktaschel@cs.ucl.ac.uk} \\
 }

\date{}

\begin{document}
\maketitle
\begin{abstract}
Natural Language Inference (NLI) datasets contain annotation artefacts resulting in spurious correlations between the natural language utterances and their respective entailment classes.
These artefacts are exploited by neural networks even when only considering the hypothesis and ignoring the premise, leading to unwanted biases.
\citet{BelinkovAdversarialRemoval} proposed tackling this problem via adversarial training, but this can lead to learned sentence representations that still suffer from the same biases.
We show that the bias can be reduced in the sentence representations by using an ensemble of adversaries, encouraging the model to jointly decrease the accuracy of these different adversaries while fitting the data. This approach produces more robust NLI models, outperforming previous de-biasing efforts when generalised to 12 other NLI datasets~\citep{BelinkovPremiseGranted,NewBelinkovPaper}.
In addition, we find that the optimal number of adversarial classifiers depends on the dimensionality of the sentence representations, with larger sentence representations being more difficult to de-bias while benefiting from using a greater number of adversaries.
\end{abstract}

\section{Introduction}

NLI datasets are known to contain artefacts associated with their human annotation processes~\citep{AnnotationArtifacts}.
Neural models are particularly prone to picking up on artefacts, relying on these biases and spurious correlations rather than acquiring a true understanding of the task.
Because these artefacts are often dataset specific~\citep{PoliakEtAl,Tsuchiya}, models that rely on these artefacts consequently generalise poorly when tested on other datasets~\citep{BelinkovPremiseGranted}. 

One way to alleviate this problem is via \emph{adversarial training}: the task classifier and an adversarial classifier jointly share an encoder, with the adversarial classifier trained to produce the correct predictions by analysing the artefacts in the training data. The encoder optimises the training objective while also reducing the performance of the adversarial classifier.
In this context, adversarial training aims to produce sentence representations that 
do not incorporate information about the artefacts (or \emph{bias}) in the data, resulting in less biased models that generalise better.

Previous studies show that adversarial training is associated with better generalisation performance across other datasets, although there are concerns that the biases are not removed from the model sentence representations, with classifiers able to relearn such biases from the representations after these are frozen~\citep{BelinkovAdversarialRemoval, ElazarAndGoldberg}. It is therefore unclear whether any improvements are as a result of the de-biasing, and whether actually removing these biases from the model representations will further improve generalisation. 
We focus our effort on this discrepancy and argue that, in order to show de-biasing is effective, improvements in performance should also correspond to an observed reduction of the bias in the model representations, therefore creating representations that generalise better to other data. 

In this paper we show that NLI models can avoid learning from the hypothesis-only bias, using an \emph{ensemble of adversarial classifiers} to prevent the bias being relearnt from a model's representations. Furthermore, we show that the more bias is removed from the model representations, the better these models generalise to other NLI datasets. Removing the bias from the representations proves to be a highly effective strategy, producing more robust NLI models that outperform previous de-biasing efforts when tested on 12 different NLI datasets~\citep{BelinkovPremiseGranted,NewBelinkovPaper}. In addition, we show that the ability to de-bias a sentence representation depends on its dimensionality, with large sentence representations being harder to de-bias and requiring more adversarial classifiers during training. 

In summary, this paper makes the following core contributions:
\begin{inparaenum}[\itshape i)\upshape]
%
\item We investigate whether using an ensemble of adversarial classifiers can remove the hypothesis-only bias within NLI models. For large enough dimensions, this method achieves a statistically significant reduction of the bias. 
\item We test whether removing more of the bias improves how well the model generalises. Our method improves model accuracy across 12 NLI datasets and outperforms previous research~\citep{BelinkovPremiseGranted,NewBelinkovPaper}.
\item We inspect the optimal number of adversaries to use depending on the dimensionality of the sentence representations. We find that as this dimensionality is increased, more adversaries are required to de-bias a model.
\item We compare the effect of adversarial training with a linear classifier to using a non-linear multi-layer perceptron as the adversary, showing that using a more complex adversarial classifier is not always beneficial. Instead, the best choice of adversary depends on the classifier being used to relearn the bias.
\footnote{\url{https://github.com/joestacey/robust-nli}}

%
\end{inparaenum}

\section{Related Work}
\paragraph{The Hypothesis-Only Bias}
\citet{AnnotationArtifacts} and \citet{Tsuchiya} demonstrate how models can predict the class within the SNLI dataset when only processing the hypothesis, reaching accuracy scores as high as twice the majority baseline (67\% vs. 34\%). This is possible due to hypothesis-only biases, such as the observation that negation words (``no'' or ``never'') are more commonly used in contradicting hypotheses~\citep{AnnotationArtifacts,PoliakEtAl}. The hypothesis sentence length is another example of an artefact that models can learn from, with entailment hypotheses being, on average, shorter than either contradiction or neutral hypotheses~\citep{AnnotationArtifacts}.
\citet{Tsuchiya} show that the hypothesis-only bias predictions are significantly better than the majority baseline for SNLI, although this is not the case for the SICK dataset~\citep{DBLP:conf/lrec/MarelliMBBBZ14}.
\citet{PoliakEtAl} find that human-elicited datasets such as SNLI and MultiNLI have the largest hypothesis-only bias.
As a result, our paper focuses on removing the hypothesis-only bias from SNLI, the dataset with the largest hypothesis-only bias reported by \citet{PoliakEtAl}. This bias is also dataset specific, with \citet{BelinkovPremiseGranted} finding that only MultiNLI shares some of the same hypothesis-only bias as the SNLI dataset.
\paragraph{Generalisation to Other Datasets}
\citet{SNLI} and \citet{UnderstandingThroughInference} show that models trained on the SNLI and MultiNLI datasets do not necessarily learn good representations for other NLI datasets, such as SICK. Analogous results were also reported by \citet{DBLP:journals/corr/abs-1810-09774} for more complex models.
\citet{AnnotationArtifacts} and \citet{Tsuchiya} identify how NLI models perform worse on \emph{hard} examples, which are defined as the examples that a hypothesis-only model has misclassified. This suggests that the success of NLI models may be overstated, with models relying on artefacts in their training data to achieve high performance  \citep{AnnotationArtifacts}. Our paper will assess whether NLI models that no longer learn from the hypothesis-only bias can still retain this high level of accuracy.
\paragraph{Biases and Artefacts}
SNLI and MultiNLI are not the only datasets that suffer from the presence of annotation artefacts and biases.
In the past, machine reading datasets were also found to contain syntactic clues that were giving away the correct prediction~\citep{DBLP:conf/mlcw/VanderwendeD05,DBLP:conf/naacl/SnowVM06}.
For instance, \citet{DBLP:conf/emnlp/KaushikL18} show that, in several reading comprehension datasets such as bAbI~\citep{DBLP:journals/corr/WestonBCM15} and Children’s Books Test~\citep{DBLP:journals/corr/HillBCW15}, it is  possible to get non-trivial results by considering only the last passage of the paragraph.
In visual question answering datasets, several studies find it is often possible to answer the question without looking at the corresponding image~\citep{DBLP:conf/cvpr/ZhangGSBP16,DBLP:conf/cvpr/KafleK16,DBLP:conf/cvpr/GoyalKSBP17,DBLP:conf/cvpr/AgrawalBPK18}.
Similarly, for the ROCStories corpus~\citep{DBLP:conf/naacl/MostafazadehCHP16}, \citet{DBLP:conf/conll/SchwartzSKZCS17} and \citet{DBLP:conf/acl/CaiTG17} show it is possible to achieve non-trivial prediction accuracy by only considering candidate endings and without taking the stories in account.
The de-biasing approach introduced in this paper could be applied in any of these situations where a model involves a classifier based on latent representations. 
\paragraph{Learning Robust Models}
Neural models are known to be vulnerable to so-called \emph{adversarial examples}, \ie{} instances explicitly crafted by an adversary to cause the model to make a mistake~\citep{IntriquingProperties}.
Most recent work focuses on simple semantic-invariant transformations, showing that neural models can be overly sensitive to small modifications of the inputs and paraphrasing.
For instance, \citet{DBLP:conf/acl/SinghGR18} use a set of simple syntactic changes, such as replacing \emph{What is} with \emph{What's}.
Other semantics-preserving perturbations include typos~\citep{DBLP:conf/cvpr/HosseiniXP17}, the addition of distracting sentences~\citep{DBLP:conf/naacl/WangB18, JiaAndLiang}, character-level perturbations~\citep{DBLP:conf/acl/EbrahimiRLD18}, and paraphrasing~\citep{DBLP:conf/naacl/IyyerWGZ18}.
\citet{PasqualeLogicPaper} propose searching for violations of constraints, such as the symmetry of \emph{contradiction} and transitivity of \emph{entailment}, for identifying where NLI models make mistakes and then creating more robust models by training on these adversarial examples. 
Alternatively, \citet{DontTakeEasyWayOut}, \citet{He2019UnlearnDB} and \citet{NewBelinkovPaper} create naive models that make predictions based on known dataset biases, and then train robust models in an ensemble with the naive models to focus on other patterns in the data that generalise better.

\paragraph{Adversarial Training}

Another procedure for creating more robust models is through adversarial training with latent representations, with a classifier trained to learn the bias from the model sentence representations which in turn update to reduce the performance of the bias classifier~\citep{DBLP:conf/naacl/WangSL19}. For example, \citet{GaninPaperVision} use adversarial training to improve domain adaption, allowing models to learn features helpful for the model task but which are also invariant with respect to changes in the domain.
This was achieved by jointly training two models, one to predict the class label and one to predict the domain, and then regularising the former model to decrease the accuracy of the latter via gradient reversal.
\citet{BelinkovAdversarialRemoval} use adversarial training to remove the hypothesis-only bias from models trained on SNLI.
While this approach produced models that generalised better to other datasets, these same models show degraded performance on SNLI-hard \citep{BelinkovPremiseGranted}, which is supposedly the ideal dataset to test for generalisation as it resembles SNLI the most in terms of domain and style while lacking the examples with the largest bias \cite{AnnotationArtifacts}. Moreover, the bias is not removed from the model sentence representations and can be almost fully recovered if these representations stop updating \cite{BelinkovAdversarialRemoval}. It is therefore unclear whether any improvements are caused by de-biasing, or are instead a result of perturbations from the adversarial training procedure.
Similarly, \citet{ElazarAndGoldberg} find that adversarial training and gradient reversal does not remove demographic information such as age or gender, with this information still present in the de-biased sentence representations.
Here, we propose using an ensemble of multiple adversaries to avoid the hypothesis-only bias, significantly reducing the bias stored within a model's representations and outperforming previous de-biasing research efforts when testing model generalisation.
\paragraph{Using Model Ensembles}

\citet{StrengthInNumbersOLD} find that using ensembles of models is a better use of computational budget when training from adversarial examples compared to using a larger model with more parameters.
\citet{ElazarAndGoldberg} show that using an ensemble of up to 5 adversarial classifiers helped remove demographic information contained within Twitter messages, however beyond this they were not able to re-learn the main task.
\citet{NewBelinkovPaper} provide further support for using model ensembles after implementing a Product of Experts approach with multiple hypothesis-only models, producing more robust NLI models despite not attempting to remove the bias from the underlying representations. We compare our results with \citet{NewBelinkovPaper} to quantify the benefits of removing the bias from the sentence representations.
\section{Ensemble Adversarial Training}
We follow an adversarial training approach for reducing the hypothesis-only bias contained within the sentence representations.
Specifically, we generalise the adversarial training framework proposed by \citet{BelinkovPremiseGranted} to make use of \emph{multiple adversaries}: $n$ 
hypothesis-only adversaries
are jointly trained for predicting the relationship between the premise and hypothesis given only the representation of the hypothesis from the sentence encoder.
At the same time, the sentence encoder together with an hypothesis-premise model are jointly trained to fit the training data, while decreasing the accuracy of these adversaries.
Formally, given a hypothesis $\mathbf{h}$ and a premise $\mathbf{p}$, the predictions of the hypothesis-premise model $\hat{y}$ and the $i$-th hypothesis-only adversary $\hat{y}_{a_{i}}$ can be formalised as follows:
\begin{equation*}
\begin{aligned}
\mathbf{e}_{h} & = \encoder_{\params_{e}}(\mathbf{h}), \quad \mathbf{e}_{h} \in \mathbb{R}^{k} \\
\mathbf{e}_{p} & = \encoder_{\params_{e}}(\mathbf{p}), \quad \mathbf{e}_{p} \in \mathbb{R}^{k} \\
\hat{y} & = \MLP_{\params_{c}}\left(\left[ \mathbf{e}_{h} ; \mathbf{e}_{p} ; \mathbf{e}_{h} - \mathbf{e}_{p} ; \mathbf{e}_{h} \odot \mathbf{e}_{p} \right]\right) \\
\hat{y}_{a_{i}} & = \MLP_{\params_{a_{i}}}\left(\mathbf{e}_{h}\right),
\end{aligned}
\end{equation*}
\noindent where $\hat{y}, \hat{y}_{a_{i}} \in \mathbb{R}^{3}$ are the probability distributions produced for the three NLI classes, \ie{} \emph{entailment}, \emph{contradiction}, and \emph{neutral}, and $\params_{e}, \params_{c}, \params_{a_{i}}$ respectively denote the parameters of the encoder, the hypothesis-premise model, and the $i$-th hypothesis-only adversary.
The adversarial training procedure can be formalised as optimising the following \emph{minimax objective}:
\begin{equation} \label{eq:loss}
\begin{aligned}
\min_{\params_{e}, \params_{c}} \max_{\params_{a}} \sum_{\langle \mathbf{h}, \mathbf{p}, y \rangle \in \mathcal{D}} & (1 - \lambda) \mathcal{L}_{ce}(y, \hat{y}) \\ 
- & \frac{\lambda}{n} \sum_{i = 1}^{n} \mathcal{L}_{ce}(y, \hat{y}_{a_{i}}),
\end{aligned}
\end{equation}
\noindent where $\mathcal{D}$ is a dataset, and $\mathcal{L}_{ce}$ denotes the cross-entropy loss~\citep{DBLP:books/daglib/0040158}, and $n \in \mathbb{N}_{+}$ is the number of adversaries.
The hyperparameter $\lambda \in \left[0, 1 \right]$ denotes the trade-off between the losses of the hypothesis-premise model and the hypothesis-only adversaries.
Similarly to \citet{BelinkovPremiseGranted}, we optimise the minimax objective in \cref{eq:loss} using gradient reversal~\citep{GaninPaperVision}, which leads to an optimisation procedure equivalent to the popular gradient descent ascent algorithm~\citep{DBLP:journals/corr/abs-1906-00331}. The gradient reversal multiplies the gradient from the adversarial classifiers by a negative constant, training the encoder to reduce the performance of these classifiers.
%

%
%

%
To test the impact of using multiple adversarial classifiers when changing the dimensionality of the representations, we train with $\{1, 5, 10, 20\}$ bias classifiers for $\{ 256, 512, 1024, 2048 \}$ dimensional sentence representations.
The learned sentence representation is then frozen, and $20$ adversarial classifiers are randomly reinitialised before they attempt to re-learn the hypothesis-only bias from the frozen de-biased sentence representation. The maximum accuracy from across the $20$ adversarial classifiers is then reported after trying to remove the bias, showing the maximum bias that can still be learnt from the representation.
The ability of adversarially trained models to de-bias sentence representations is tested across a range of $\lambda$ hyper-parameters $\{0.001, 0.01, 0.1, 0.2, 0.3, ... 0.8, 0.9, 0.99, 0.999\}$. This shows whether any improvement is due to the choice of $\lambda$, or whether there is an improvement regardless. 
\paragraph{Model Architecture}
Following the same experimental set-up as \citet{BelinkovPremiseGranted} and \citet{PoliakEtAl}, we use an InferSent model~\citep{InferSent} with pretrained GloVe 300-dimensional word embeddings. The InferSent model architecture consists of a
Long Short-Term Memory network~\citep[LSTM,][]{DBLP:journals/neco/HochreiterS97} encoder which creates a $2048$ dimensional sentence representation.
\begin{figure*}[t]
    \includegraphics[width=\textwidth]{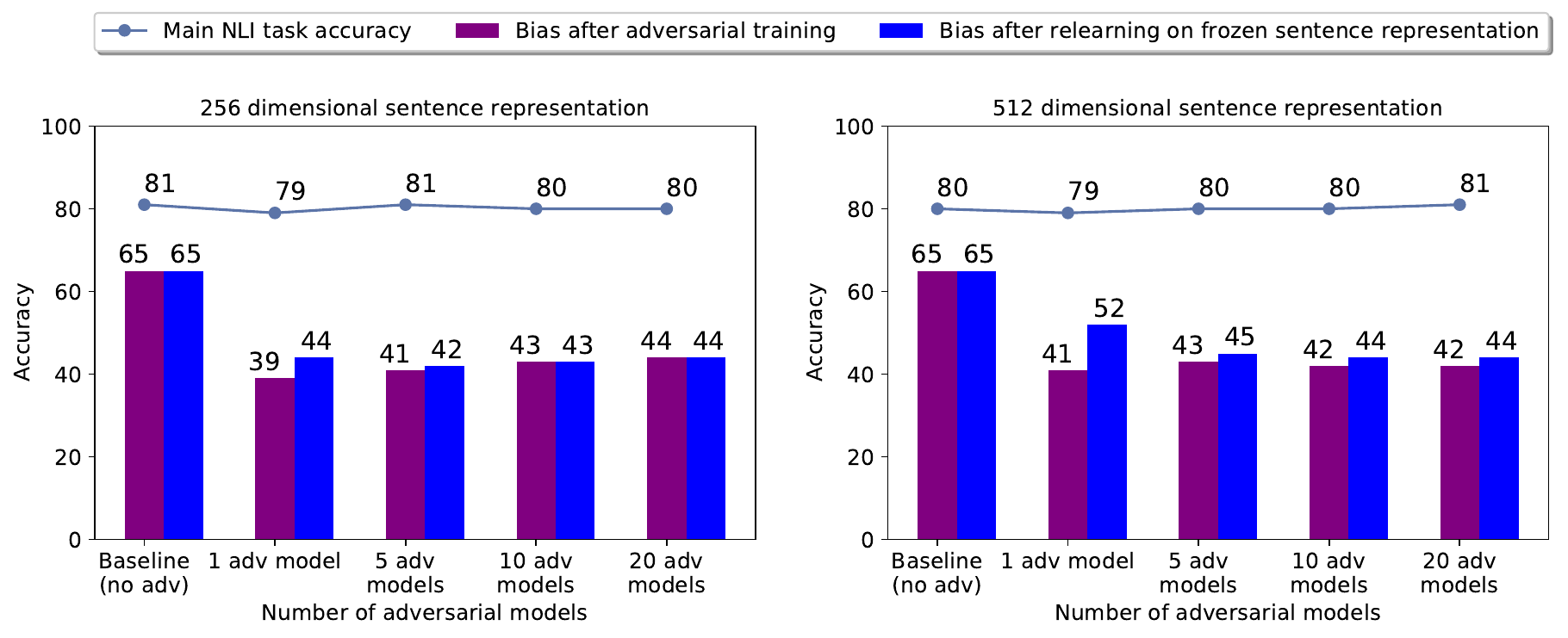}
    \includegraphics[width=\textwidth]{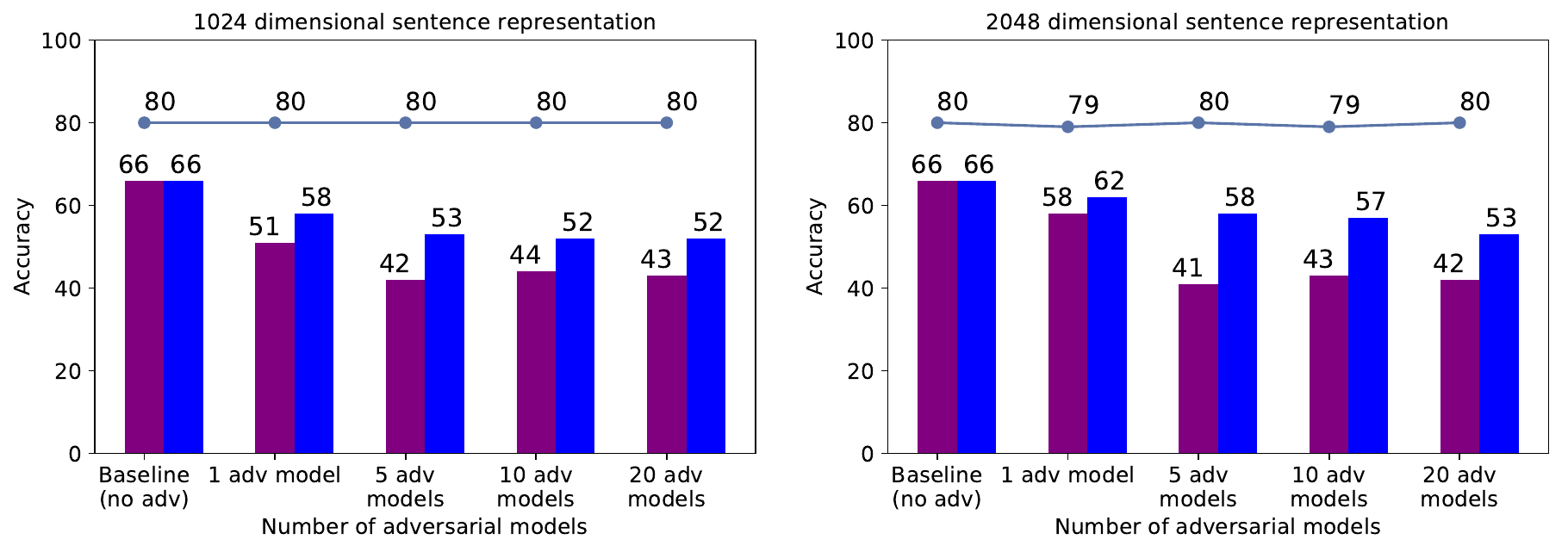}
    \caption{The maximum bias classifier accuracy after the bias is re-learnt from the frozen de-biased representations (in blue), compared to the accuracy of an independent bias classifier (not an adversary) at the end of the adversarial training (in purple). The main NLI task accuracy is also displayed.}
    \label{full_results}
\end{figure*}
\subsection{Significance Testing}
We perform statistical testing to assess whether the differences between using one or five adversarial classifiers is significant. This involves repeating the experiments for both one and five adversarial classifiers with ten different random seeds. For each experiment, the de-biasing is performed before a classifier attempts to learn the bias again from the frozen sentence representations.
We use bootstrapping hypothesis testing~\citep{BootstrapBook} to test the statistical significance by comparing the means from the two samples. We also provide p-values from a Mann Whitney U-test~\citep{MannWhitneyUTest}. The bootstrapping considers the null hypothesis that there is no difference between the mean bias re-learnt from using five adversarial classifiers compared to just using one adversarial classifier. In addition, we use a Bonferroni correction factor~\citep{BonferroniCorrection} of four when evaluating the p-values, taking into account multiple hypothesis testing across each different dimension. P-values smaller than 0.05 are considered significant.

\subsection{Using Deeper Adversaries}

We also investigate using a multi-layer perceptron as a more complex adversarial classifier to understand whether the bias that can be re-learnt depends on the type of classifier used. The experiments are repeated using non-linear classifiers instead of linear classifiers, both during the adversarial training and also afterwards when the classifiers try to re-learn the biases from the frozen representations. In addition to testing $2048$ dimensions using ten adversaries, a $512$ dimensional representation is also tested using a smaller number of adversaries (five).

We perform the experiments with three scenarios:
%
\begin{inparaenum}[\itshape i)\upshape]
\item Using the non-linear classifiers during the adversarial training, but not afterwards, and instead trying to re-learn the bias with a linear classifier.
\item Using linear classifiers during the adversarial training but then non-linear classifiers are used to try to re-learn the biases after the sentence representation is frozen.
\item Finally, non-linear classifiers are used both during adversarial training and afterwards when trying to re-learn the biases from the frozen sentence representation.
%
\end{inparaenum}
The non-linear multi-layer perceptron classifier consists of three linear layers, and two non-linear layers using $\tanh$.
\subsection{Evaluating De-biased Sentence Encoders}

After training the models on SNLI with adversarial training, we test these de-biased models on a range of different datasets to see whether they generalise better. The performance of the de-biased models is compared to a baseline model trained on SNLI where no adversarial training has been performed. 

By using different random seeds, we compare ten baseline SNLI-trained models with models using one adversary and 20 adversaries, with each of these models tested on SNLI-hard. We perform bootstrap hypothesis testing to understand whether there is a significant difference between using one adversary and the baseline models with no adversarial training. We have repeated this hypothesis testing to compare models de-biased using 20 adversaries to the baseline models.
Additionally, we evaluate the de-biased models on 12 different datasets to understand whether models trained with an ensemble of adversaries perform better than the baseline and models trained with one adversary. 
The datasets in these experiments are the same datasets tested by \citet{BelinkovPremiseGranted}: ADD-ONE-RTE~\citep{DBLP:conf/acl/PavlickC16},  GLUE~\citep{DBLP:conf/emnlp/WangSMHLB18}, JOCI~\citep{DBLP:journals/tacl/ZhangRDD17}, MNLI~\citep{UnderstandingThroughInference}, MPE~\citep{DBLP:conf/ijcnlp/LaiBH17}, SCITAIL~\citep{DBLP:conf/aaai/KhotSC18}, SICK~\citep{DBLP:conf/lrec/MarelliMBBBZ14}, SNLI-hard~\citep{AnnotationArtifacts}, and three datasets recast by \citet{white-etal-2017-inference}:
DPR~\citep{DBLP:conf/emnlp/RahmanN12}, FN+~\citep{DBLP:conf/acl/PavlickWRCDD15} and SPR~\citep{DBLP:journals/tacl/ReisingerRFHRD15}.

While the previous results in this paper use an LSTM encoder, a bidirectional LSTM has been used when testing other datasets to ensure the experiments are a like-for-like comparison with \citet{BelinkovPremiseGranted}.
We select the hyper-parameters that yield the highest accuracy on a validation set, in line with the experiments conducted by \citet{BelinkovPremiseGranted}. Finally, the model results are compared to the Product of Experts (PoE) de-biasing approach proposed by \citet{NewBelinkovPaper}.

\begin{table}[t]
\begin{center}
\begin{tabular}{rccccc}
\toprule
& \multicolumn{5}{c}{\bf Number of adversaries}\\
Dimensions & {\bf 0} & {\bf 1} & {\bf 5} & {\bf 10} & {\bf 20} \\
\midrule
256 & 65 & 44 & \textbf{42} & 43 & 44 \\
512 & 65 & 52 & 45 & \textbf{44} & \textbf{44} \\
1,028 & 66 & 58 & 53 & \textbf{52} & \textbf{52} \\
2,048 & 66 & 62 & 58 & 57 & \textbf{53} \\
\bottomrule
\end{tabular}
\end{center}
\caption{Maximum accuracy (\%) from 20 bias classifiers when re-learning the hypothesis-only bias from the frozen de-biased sentence representation. The lowest accuracy figure for each dimension is highlighted, after testing 0, 1, 5, 10 and 20 adversaries.}
\label{summary_results}
\end{table}

\section{Results}

We use an ensemble of multiple adversarial classifiers during model training to understand whether it is possible to reduce the bias within the model sentence representations. Our results show that training a model with an ensemble of adversaries does reduce the model bias, doing so across each dimensionality of sentence representations tested. Moreover, more adversaries are required for de-biasing larger dimensional sentence representations.

When using just one adversarial classifier for a 2,048 dimensional sentence representation, after the representation was frozen and the bias classifiers had a chance to re-learn the hypothesis-only bias, the accuracy of the bias classifiers increased to 62\% (see \cref{summary_results}). This result mirrors the findings of \citet{BelinkovAdversarialRemoval}, where using one adversarial classifier does not remove the bias from the model sentence representations. We additionally show that an independent bias classifier can still reach 58\% accuracy at the end of the adversarial training, without needing to freeze the representation in order to relearn the bias (see \cref{full_results}).

For 2,048 dimensional sentence representations, as the number of adversaries are increased up to 20, less bias can be found in the resulting de-biased sentence representation. When the number of adversarial classifiers are increased from 1 to 20, the accuracy of the hypothesis-only bias classifiers reduces from 62\% to 53\% (see \cref{summary_results}).

The higher the dimensionality of the sentence representation, the more difficult it is to remove its bias. Additionally, the optimal number of adversarial classifiers depends on the dimensionality of the representations, with more adversaries required for higher dimensions. For 2,048 dimensions this is 20 adversaries, while for 256 dimensions this reduces to 5 adversaries (see \cref{summary_results}). For 256, 512 and 1,028 dimensions, the improvements plateau after a set number of adversaries, and therefore scaling up the number of adversaries beyond 20 is unlikely to lead to further improvements.

\begin{figure}[t]
    \includegraphics[width=\columnwidth]{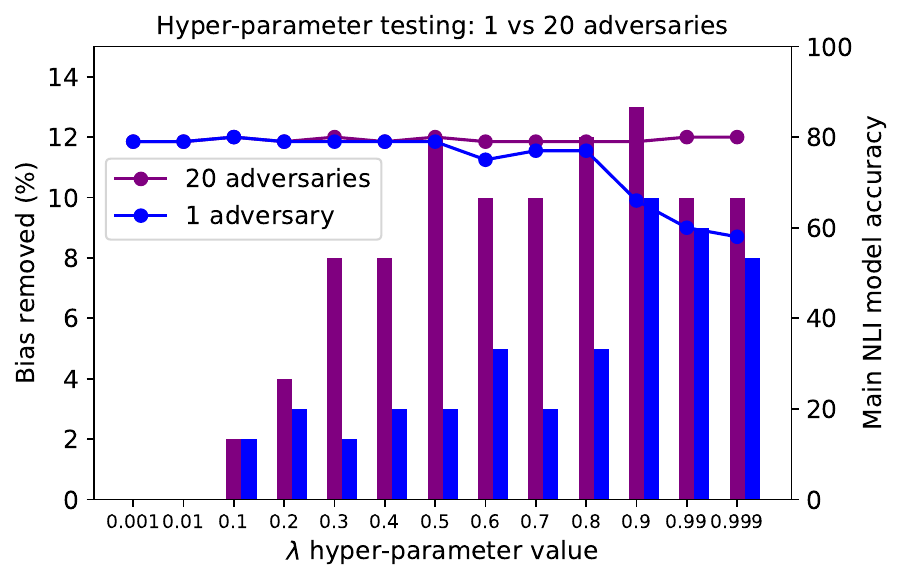}
    \caption{The fall in accuracy of hypothesis-only classifiers when using 1 or 20 adversaries to remove the hypothesis-only bias (compared to a baseline with no adversaries). This is shown alongside the overall accuracy of the de-biased NLI models.} \label{hyper_param_testing}
\end{figure}

The improvements in de-biasing when using an ensemble of adversaries is consistent across different $\lambda$ hyper-parameter values: when comparing a model trained with one adversary to a model trained with 20 adversaries, the model trained with 20 consistently removes more bias. Using a single adversary does reduce the bias for $\lambda$ values of 0.9 and above, but at the expense of the overall model accuracy which reduces dramatically (\cref{hyper_param_testing}). This suggests that valuable information about the hypothesis is being removed instead of just the bias. Interestingly, when using 20 adversaries, even for the largest values of $\lambda$, the overall model accuracy does not start to decrease.

%

\subsection{Evaluating Multiple Adversaries}

We applied statistical testing to understand whether the improvements seen using an ensemble of adversarial classifiers is statistically significant. For sentence representations with 2,048, 1,024 or 512 dimensions this is a statistically significant result. Although the results are not statistically significant for a smaller 256 dimensional representation.

\begin{figure}[t]
    \includegraphics[width=\columnwidth]{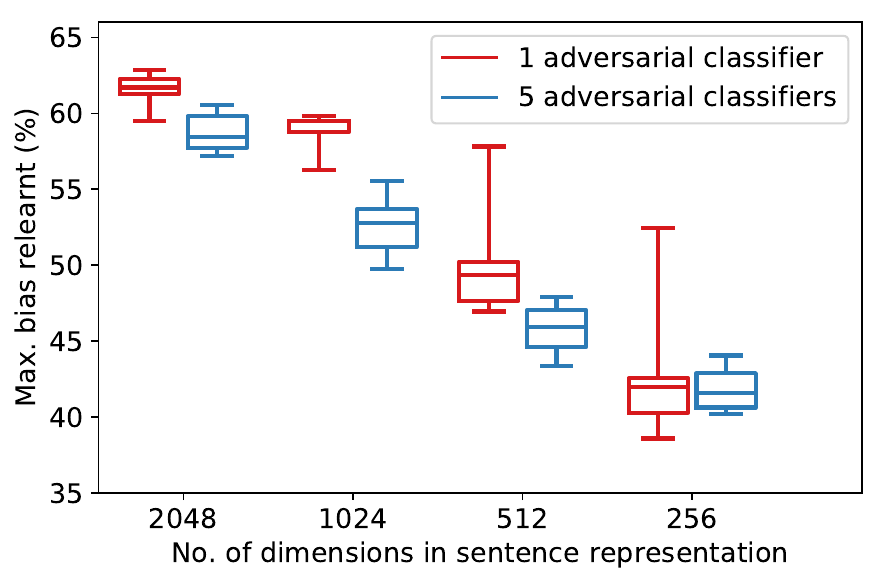}
    \caption{Maximum accuracy scores of the bias classifiers when they are retrained on de-biased sentence representations for each of the experiments tested. Ten experiments were performed for each condition, using one or five adversaries for each dimension.}
    \label{stat_testing_results}
\end{figure}

For 2,048, 1,024 and 512 dimensional sentence representations, the statistical testing provides p-values smaller than 0.05. The null hypothesis is therefore rejected in these cases, with the alternative hypothesis stating that using five adversaries reduces the mean bias re-learnt from the sentence representations compared to using just one adversarial classifier (see \cref{significance_testing}). \cref{stat_testing_results} displays these results in a boxplot diagram.

%
%

\begin{table*}[t]
\begin{center}
\begin{tabular}{rcccc}
\toprule
& \multicolumn{2}{c}{\bf P-Value} & \multicolumn{2}{c}{\bf Accuracy of bias classifier} \\
%
\midrule
{\bf Dim.} & {\bf Mann-Whitney} & {\bf Bootstrapping} & {\bf Mean (Median), 1 Adv.} & {\bf Mean (Median), 5 Adv.} \\
\midrule
256 & 1 & 1 & 42.4 (42.0) & 41.8 (41.6) \\
512 & 0.0009* & \textless 0.0001* & 49.8 (49.4) & \textbf{45.8} (45.9) \\
1,028 & 0.0008* & \textless 0.0001* & 58.9 (59.5) & \textbf{52.6} (52.8) \\
2,048 & 0.0005* & 0.0105* & 61.6 (61.7) & \textbf{58.7 }(58.4) \\
\bottomrule
\end{tabular}
\end{center}
\caption{p-values after performing Bootstrapping and Mann-Whitney hypothesis tests, using a Bonferroni correction factor of 4. * indicates a statistically significant result with a p-value below 0.05. Highlighted values indicate that the mean is significantly smaller than its comparison mean value, using the bootstrapping p-values.}
\label{significance_testing}
\end{table*}

\begin{table}[t]
\begin{center}
\begin{tabular}{rcc}
\toprule
\textbf{Adversary type} & \textbf{Class. type} & \textbf{Accuracy} \\
\midrule
\textit{2,048 dim (10 adv)} & & \\
\midrule
Linear & Linear & 56\\
Linear & Non-linear & 66 \\
Non-Linear & Linear & 61 \\
Non-Linear & Non-linear & 62 \\
\midrule
\textit{512 dim (5 adv)} & & \\
\midrule
Linear & Linear & 44 \\
Linear & Non-linear & 66 \\
Non-Linear & Linear & 55 \\
Non-Linear & Non-linear & 60 \\
\bottomrule
\end{tabular}

\end{center}
\caption{Accuracy of bias classifiers when relearning the bias after using either a linear or non-linear adversary, when the classifier used to re-learn the bias (\textit{class. type}) is also either linear or nonlinear.}
\label{non-linear_results}
\end{table}

\begin{table}[t]
\begin{center}
\resizebox{\columnwidth}{!}{
\begin{tabular}{rrrrr}
\toprule
{\bf Dataset} & {\bf Baseline} & {\bf PoE} & {\bf 1 Adv.} & {\bf Ens.}  \\
\midrule
AOR & 61.24 & +1.8 & -2.3 & +1.3 \\
DPR & 46.30 & +0 & +2.9 & +0.9 \\
FN+ & 38.43 & +0.7 & +6.8 & +12.2 \\
GLUE & 43.12 & +0.4 & +0.6 & -1.0 \\
JOCI & 40.77 & +1.1 & +0.8 & +1.6 \\
MNLI Match. & 53.38 & -0.3 & +0.8 & +0.8 \\
MNLI Mism. & 52.91 & +0.8 & -0.5 & -0.1 \\
MPE & 57.30 & +1.2 & -0.4 & +0.5 \\
SCIT & 47.98 & -1.1 & +0.5 & +0.5 \\
SICK & 50.61 & +1.7 & 0.4 & -0.4 \\
SNLI Hard & 65.72 & +4.2 & +1.2 & +1.7 \\
SNLI & 83.29 & -3.9 & +0.2 & +0.8 \\
SPRL & 30.35 & +9.2 & +10.3 & +13.5 \\
\midrule
{\bf Average} & & {+1.2} & +1.6 & \emph{\bf +2.5} \\
\bottomrule
\end{tabular}
}
\end{center}
\caption{Accuracy of the de-biased models when tested on 12 different NLI datasets, comparing models trained with one adversary (1 Adv.) as per \citet{BelinkovPremiseGranted}, to models trained with an ensemble of adversaries (Ens.). The Product of Experts (PoE) approach proposed by \citet{NewBelinkovPaper} is also included.}
\label{datasets}
\end{table}

\subsection{Using Deeper Adversaries}

To investigate the impact of changing the strength of the adversary, multi-layer perceptrons are used during model training as the adversarial classifiers. The results show that more complex multi-layer perceptrons do not always perform better, and that the best choice of adversary depends on the type of classifier used  to relearn the bias.

When a non-linear model is used to re-learn the bias from the frozen sentence representation, less bias can be recovered if a non-linear model was used as the adversarial classifier during training instead of a linear adversarial classifier (see \cref{non-linear_results}). Therefore, when using a more complex classifier to re-learn the bias, a model of at least the same complexity should be used in the adversarial training to remove these biases. If a linear classifier is used as the adversary, a non-linear classifier can find more bias when learning from the de-biased representation than a linear classifier can.

The results also show that if a linear model is being used to re-learn the bias, then using a linear model as the adversary instead of a multi-layer perceptron reduces the amount of bias that can be recovered (see \cref{non-linear_results}). This could suggest that the best approach is to use the same type of classifier for both the adversarial model and the classifier used to re-learn the bias. However, more adversarial classifiers may be required when using non-linear classifiers as adversaries, and therefore more experimentation is required to test this hypothesis.

The classifier chosen to relearn the bias will depend on the model that is being de-biased. If a classifier cannot relearn the bias from the sentence representations, this is a guarantee that a model using the de-biased representations and this classifier will not be influenced by the bias.

\subsection{Evaluating De-biased Encoders}

The models trained with an ensemble of adversaries are applied to 12 different NLI datasets to test whether these de-biased models generalise better than models trained with either one or no adversarial classifier. The datasets tested include SNLI-hard, where models that are no longer influenced by the hypothesis-only bias are expected to perform better. Models  trained using an ensemble of adversaries performed better across most of these datasets, including SNLI-hard where there is a statistically significant improvement compared to a baseline model with no adversarial training.

Models trained with one adversary were not significantly better than a baseline model when tested on SNLI-hard (1.1\% improvement, corresponding to a p-value of 0.07).
On the other hand, there is a statistically significant improvement when using an ensemble of 20 adversarial classifiers (achieving a 1.6\% improvement with a p-value of 0.015). As a result, we accept the alternative hypothesis that models trained with 20 adversaries have a higher mean accuracy than the baseline models.

Across 8 of the 13 datasets analysed, models trained with an ensemble of 20 adversarial classifiers performed better than when using only one adversarial classifier (see \cref{datasets}). For three of the remaining datasets, the performance was the same between using one adversary and 20 adversaries. The performance when using an ensemble of adversaries was on average 0.9 points higher than when using one adversary, which in turn outperformed the baseline by 1.6 points. The ensemble of adversaries also outperforms the Product of Experts approach proposed by  \citet{NewBelinkovPaper}.

\subsection{Discussion}

We find that the higher the dimensionality of the representations, the less effective a single adversary is at removing the bias, with the de-biasing also dependant on the strength of the classifier. These differences explain why past research has found that biases can remain hidden within the model representations~\citep{ElazarAndGoldberg, BelinkovAdversarialRemoval}, with \citet{ElazarAndGoldberg} using a model with a non-linear classifier while \citet{BelinkovAdversarialRemoval} use a single adversary with 2,048 dimensional sentence representations.

\section{Conclusions}
We set out to prevent NLI models learning from the hypothesis-only bias by using an ensemble of adversarial classifiers. Our method produced sentence representations with significantly less bias, and these more robust models generalised better to 12 different NLI datasets, improving over previous approaches in the literature~\citep{BelinkovAdversarialRemoval,NewBelinkovPaper}. 

The higher the dimensionality of the sentence representations, the harder it is to de-bias these representations and the higher the optimal number of adversarial classifiers appears to be.
Furthermore, the models trained with an ensemble of adversaries also performed better when tested on SNLI-hard compared to using only one adversarial classifier. This is the behaviour expected from de-biased models that no longer use the hypothesis-only bias to inform their predictions.

By preventing a linear classifier from learning the bias from the de-biased representations, we conclusively show that a model using such a classifier with these representations will not make decisions based on the bias. However, after implementing the adversarial training, a non-linear classifier may still be able to detect the bias in the sentence representations where linear classifiers are not able to. 
While we illustrate the conditions under which biases are removed from a linear classifier, preventing a non-linear classifier from learning the biases is more difficult and merits further experimentation.

\section*{Acknowledgements}
We would like to thank Yonatan Belinkov and Rabeeh Karimi Mahabadi for providing us with their code and hyper-parameters so that we could reproduce and review their previous work. We would also like to thank Adam Poliak for publishing a quality code base that we were able to adapt. We also thank the anonymous reviewers for their thoughtful feedback and the amazing UCL NLP group for all their encouragement along the way.
Pasquale Minervini's work is supported by the EU Horizon 2020 Research and Innovation Programme under the grant 875169.

\bibliography{bibliography}
\bibliographystyle{acl_natbib}


\end{document}